\documentclass[10pt,twocolumn,letterpaper]{article}

\usepackage{wacv}
\usepackage{times}
\usepackage{epsfig}
\usepackage{graphicx}
\usepackage{amsmath}
\usepackage{amssymb}

\usepackage{centernot}
\usepackage{stackengine}
\usepackage{makecell}
\usepackage{color,soul}
\usepackage{algorithm}
\usepackage{algpseudocode}

\wacvfinalcopy 

\def\wacvPaperID{585} 

\ifwacvfinal\fi
\begin{document}

\title{Simultaneous Detection and Removal of Dynamic Objects in Multi-view Images}

\author{Gagan Kanojia \quad\quad\quad Shanmuganathan Raman\\
Indian Institute of Technology Gandhinagar, India\\
{\small\{\tt gagan.kanojia, shanmuga\}@iitgn.ac.in}}

\maketitle
\ifwacvfinal\thispagestyle{empty}\fi

\begin{abstract}
Consider a set of images of a scene consisting of moving objects captured using a hand-held camera. In this work, we propose an algorithm which takes this set of multi-view images as input, detects the dynamic objects present in the scene, and replaces them with the static regions which are being occluded by them. The proposed algorithm scans the reference image in the row-major order at the pixel level and classifies each pixel as static or dynamic. During the scan, when a pixel is classified as dynamic, the proposed algorithm replaces that pixel value with the corresponding pixel value of the static region which is being occluded by that dynamic region. We show that we achieve artifact-free removal of dynamic objects in multi-view images of several real-world scenes. To the best of our knowledge, we propose the first method which simultaneously detects and removes the dynamic objects present in multi-view images.
\end{abstract}
\section{Introduction}
The advent of digital photography has changed the way of capturing and saving photographs. Nowadays, it is not uncommon to take multiple photographs of the same scene. While taking photographs of a scene at a public place, it is very likely to have moving objects, like people, vehicles, etc., present in the scene. Very often, it is not desirable to have them in the photographs. To deal with this problem, one can obtain masks highlighting the objects to be removed from the user in each image and then remove them using single image completion techniques \cite{criminisi2004region,drori2003fragment,iizuka2017globally,ulyanov2017deep,liu2018image}. However, there are two major problems with this approach. Firstly, it requires user input and secondly, single image completion techniques either rely on the image statistics or the model obtained by training on a large number of images. Hence, it is not necessary that the filled region will be similar to the static region which is occluded by the dynamic object. To avoid user input, one can detect the dynamic objects present in the scene using a set of photographs of the same scene and then remove them.\\
Detection of moving objects present in the scene has been in itself an active area of research for a long time now. In many applications, the moving objects hold important information and hence their detection plays a crucial part \cite{dekel2014photo,kanojia2014shot}. However, there are many applications where they are treated as noise and need to be dealt with. Previously, the detection of dynamic objects was performed on videos. The videos contain spatiotemporal information which can be exploited for this task. However, they require large memory and are computationally expensive due to a large number of frames. Recently, researchers have moved on to perform these tasks on a sparse sample of frames from videos. We call it an image sequence. Although an image sequence requires lesser memory to store and transmit and is computationally efficient, it poses certain challenges regarding finding correspondences and handling deformations and occlusions.\\
In this work, we address the problem of detection and removal of dynamic objects present in the multi-view images, simultaneously. The algorithm takes a set of multi-view images as input. Then, we pick one of the images as the reference image and the rest as the source images. The task is to simultaneously detect and remove the dynamic objects in the reference image by utilizing the information present in the source images. Our objective is to detect the dynamic objects without any user intervention and fill those regions with the static regions which are occluded by those objects. We exploit the coherency present in the natural scenes to achieve this. The proposed algorithm relies on the correspondences in the static regions which are easier to obtain in comparison to the dynamic objects. \\
{\bf Challenges.} The images of a scene captured by a group of people are not aligned. The dynamic objects can move a large distance or even leave the scene, due to which estimating optical flow for the dynamic objects is erroneous. We do not have any information regarding dynamic objects present in the source images. We do not assume that the dynamic objects in the reference image are present in all the source images. Since the dynamic objects do not obey the epipolar constraint between the pair of images, it can be exploited to find the dynamic objects \cite{dafni2017detecting}. However, it will not provide information about the static region which is occluded. These reasons make the problem of detection of the dynamic objects and simultaneously filling them with their static counterparts extremely difficult. The major contributions of the work are as follows.
\begin{enumerate}
\item We propose a novel technique which simultaneously detects and removes the dynamic objects present in the multi-view images.
\item We achieve an artifact-free transfer of the static regions from the source images to the reference image to fill the dynamic regions which are occluding them in the reference image.
\item We do not rely on the matches obtained on the dynamic objects to detect or to remove them.
\item We exploit the coherency present in the natural scenes to detect and remove the dynamic objects by filling those regions with the corresponding static regions which are occluded by them.
\end{enumerate}
The rest of the paper is organized as follows. Section \ref{Related Works} discusses the related work. Section \ref{Proposed Approach} describes the proposed approach in detail. Section \ref{Results and Discussions} discusses the results obtained using the proposed approach and their comparison with the state-of-the-art methods. Section \ref{Conclusion and Future Work} provides the conclusion and discusses the future scope of this work.
\begin{figure*}
	\centering
	\includegraphics[width=0.98\linewidth]{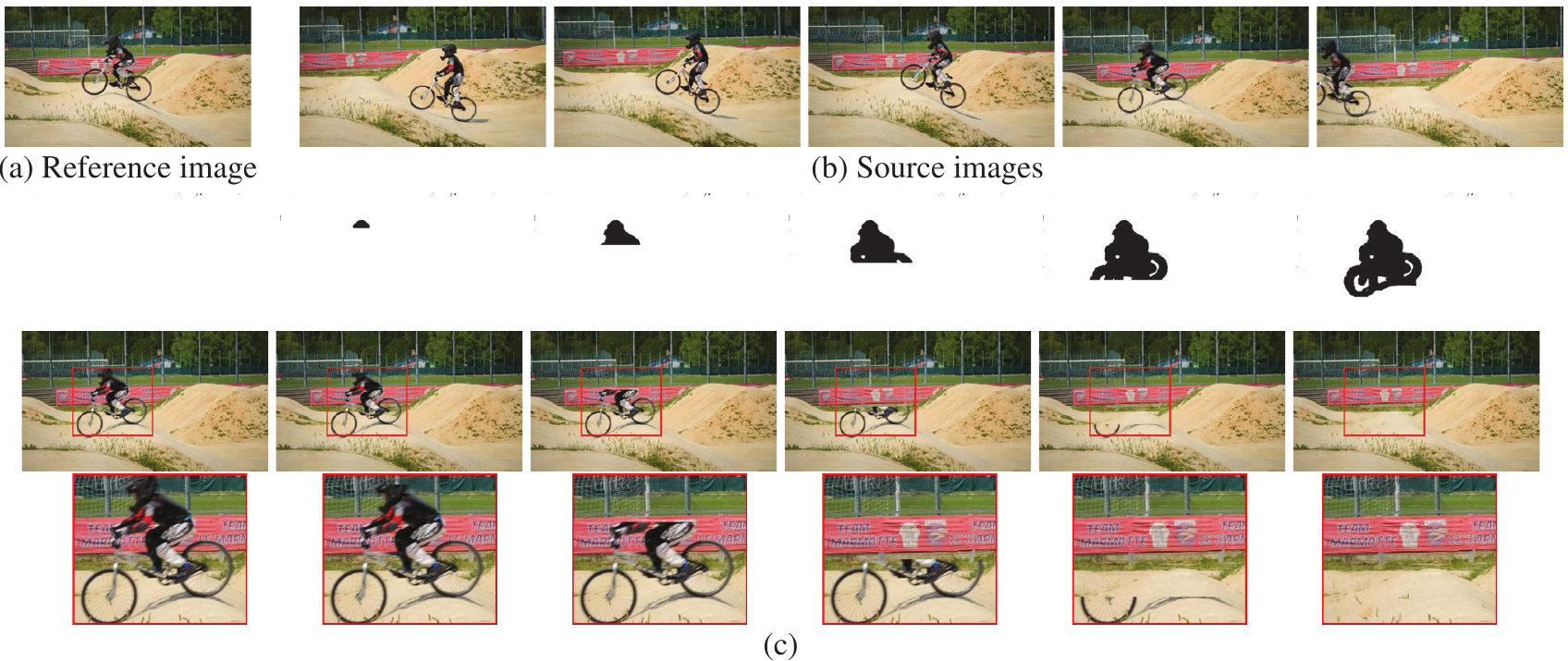}
	\caption{\textbf{Simultaneous Detection and Removal of Dynamic Objects.} The figure shows the dynamic map $\mathcal{L}$ and the updated reference image at certain intervals during the first scan (top-left to bottom right). (a) and (b) show the reference image and the source images, respectively. (c) The first and second rows show the update of the dynamic map $\mathcal{L}$ and the reference image at certain intervals of the first scan, respectively. In red border blocks, we can observe the disappearance of the dynamic object as they are being detected during the scan. The third row provides the zoomed in version of the red border blocks.}
	\label{fig: iteration}
\end{figure*}
\section{Related Works}
\label{Related Works}
\noindent{\bf Dynamic object detection in videos.} Several methods have been proposed to detect the dynamic objects in videos \cite{briassouli2007extraction,chen2013automatic,stuckler2013efficient,ochs2014segmentation,ji2014robust}. Shi and Malik proposed a moving object detection algorithm in which they treat video frames as a 3D spatiotemporal data \cite{shi1998motion}. Cremers and Soatto proposed a variational approach for segmenting the image plane into segments with parametric motion \cite{cremers2005motion}. Later, several clustering based algorithms were proposed for the task of detecting moving objects in the videos \cite{lauer2009spectral,brox2010object,lin2016diffusion}. Zhou \emph{et al.} proposed a unified framework which jointly addresses object detection and background learning using an alternating optimization \cite{zhou2013moving}. Unger \emph{et al.} showed a variational formulation for joint motion estimation and segmentation \cite{unger2012joint}. In videos, spatiotemporal information is present which can be utilized to segment the moving objects. Unlike these methods, the proposed algorithm takes a set of images of a scene as input and does not rely on the quality of matches obtained on the dynamic objects for their detection. \\
{\bf Dynamic object removal in videos.}  Patwardhan \emph{et al.} presented a framework to inpaint the missing parts in the videos \cite{patwardhan2007video,patwardhan2005video}. However, their technique is limited to the cases with either no motion of the camera or a very small camera motion. Later, the problem of filling the missing regions was posed as a global optimization problem \cite{wexler2007space,kwatra2005texture}. This helped in obtaining better globally consistent results. Recently, many methods have been proposed to deal with the camera motion by using affine transformations \cite{granados2012background,granados2012not,newson2014video}. Also, there are many methods which rely on the dense flow fields to remove the dynamic objects \cite{shiratori2006video,strobel2014flow,matsushita2006full,huang2016temporally,le2017motion,xu2017spatio,luo2018hole}. Generally, these techniques take input from the user to specify which object needs to be removed. Unlike these methods, we do not rely on the spatiotemporal information for the dynamic objects removal. Instead, we exploit the coherency present in the natural images.\\
{\bf Image inpainting in multi-view images.} Thonat \emph{et al.} proposed a method which takes a set of multi-view images and the masks of the objects which need to be removed as input and performs a multi-view consistent inpainting \cite{thonat2016multi}. Later, Philip and Drettakis introduced a plane-based multi-view inpainting technique which utilizes the local planar regions to provide more consistent multi-view inpainting results \cite{philip2018plane}. Recently, Li \emph{et al.} introduced a technique which takes an RGB-D sequence as the input to perform multi-view inpainting \cite{liu2018image}. Unlike \cite{thonat2016multi,philip2018plane}, we do not perform multi-view 3D reconstruction which itself requires handling of the dynamic objects present in the scene. We do not utilize any depth information related to the input images. Also, our objective is different from these works. Our goal is to detect the dynamic objects present in an image of the input set and fill those regions using the remaining images of the set.\\
{\bf Dynamic object detection in image sequences.} Wang \emph{et al.} proposed a method which estimates how an object has moved between a pair of images \cite{wang2015dynamic}. However, in their work, the dynamic object has to be present in both the images. Also, they rely on the point correspondences obtained on the dynamic objects. Later, Dafni \emph{et al.} proposed a method which takes a set of images of a scene consisting of dynamic objects and outputs a map highlighting the dynamic objects present in the scene \cite{dafni2017detecting}. Recently, Kanojia \emph{et al.} presented a technique which exploits image coherency to detect the dynamic objects present in a set of images of a dynamic scene \cite{kanojia2018patch}.\\
In this work, we are interested not only in the detection of dynamic objects present in a set of multi-view images of a dynamic scene, but also their removal by replacing them with the static regions which are occluded by them. Unlike \cite{kanojia2018patch}, our algorithm is iterative in nature. The changes occurring in one scan are taken foward to the next, since the reference image is updated. The way we update the reference image and the dense correspondence field during the scan to achieve the task of simultaneaous detection and removal of dynamic objects are our novel contributions.
\section{Proposed Approach}
\label{Proposed Approach}
The proposed algorithm takes a set of $n$ images of a dynamic scene captured using a hand-held camera as input. An image from the set is labeled as a reference image $I_r$ and the remaining images are labeled as source images $\{I_s\}_{s=1}^{n-1}$. Then, the algorithm scans the reference image in a row-major order at the pixel level. During the scan, at each pixel location of the reference image, it labels the pixel as static or dynamic using the information from the source images. If a pixel gets labeled as static, we move on to the next location. On the other hand, if a pixel gets labeled as dynamic, its pixel value gets updated by the corresponding pixel value of the static region which is being occluded by it. We maintain a map $\mathcal{L}:\mathbb{N}\times\mathbb{N}\rightarrow \{0,1\}$ corresponding to the reference image and keep updating it during the scans. We call it a dynamic map. Here, $0$ stands for dynamic and $1$ stands for static. First, the algorithm scans the image from top-left to bottom-right, then from bottom-right to top-left, again from top-left to bottom-right, and so on, until there is no pixel in the image which gets labeled as dynamic. The algorithm outputs an image with only static regions and a binary map highlighting the dynamic objects present in the reference image. We assume that the origin is at the top-left corner of the image and the coordinates increase as we move towards right or downwards.
\subsection{Dense Correspondences} 
\label{DenseCorr}
Since we are dealing with multi-view images, there will be deformations. In such cases, for comparison of two patches, the intensity values will not be suitable. Hence, we extract CIE Lab mean features and SIFT features \cite{lowe2004distinctive} for a patch of size $p\times p$ centered at each pixel location for all the images of the given set. We normalize the SIFT features by dividing them by the maximum of their values over all the images in the given set. Let $f_g:\mathbb{R}\times\mathbb{R}\rightarrow \mathbb{R}^{128}$ and $f_c:\mathbb{R}\times\mathbb{R}\rightarrow \mathbb{R}^{3}$ be the functions which map each pixel location of an image to 128 dimensional SIFT feature descriptor and CIE Lab mean feature descriptor, respectively.\\
We estimate dense correspondence map $\mathcal{N}_{r\rightarrow s}:\mathbb{R}\times\mathbb{R}\rightarrow\mathbb{R}\times\mathbb{R}$ from the reference image $I_r$ to each source image $I_s$, where $s= 1,2,\ldots,n-1$. Since, we want to exploit the coherency present in the scene, we have used dense flow fields \cite{huang2016temporally} for dense correspondence estimation. Herw, we do not rely on the quality of matches obtained on the dynamic objects, even incorrect matches on the dynamic objects will not affect the results. Algorithms like Full flow \cite{chen2016full} which can compute optical flow for large displacements can also be used to find the dense correspondences.\newline
We also compute a similarity map $C_s:\mathbb{R}\times \mathbb{R}\rightarrow \mathbb{R}$ for each source image $I_s$, where $s=1,2,\ldots,n-1$. The purpose of the similarity map is to quantify the quality of each match obtained by finding the dense correspondences between the reference image and the source images.
\begin{equation}
\begin{array}{ll}
C_s(\boldsymbol{x_r})= \hspace{-3mm}& \lambda_1S_e(f_c(\boldsymbol{x_r}),f_c(\boldsymbol{\hat{x}}),\sigma_c)+\\&  \lambda_2S_e(f_g(\boldsymbol{x_r}),f_g(\boldsymbol{\hat{x}}),\sigma_g)
+\lambda_3S_f(\boldsymbol{x_r},\boldsymbol{\hat{x}},\mathcal{F}_s)
\end{array}
\end{equation}
Here, $s =1,2,\ldots,n-1$, $\boldsymbol{x_r}=(\mathit{x},\mathit{y})$ is the pixel location in the reference image $I_r$, and $\boldsymbol{\hat{x}} = \mathcal{N}_{r\rightarrow s}(\boldsymbol{x_r})$ is the nearest neighbour location of $\boldsymbol{x_r}$ in $I_s$. $f_c(\boldsymbol{x})$ and $f_g(\boldsymbol{x})$ represent the CIE Lab mean feature vector and SIFT feature descriptor extracted at the pixel location $\boldsymbol{x}$ of an image, respectively. Here, $S_e(f_t(\boldsymbol{x_1}),f_t(\boldsymbol{x_2}),\sigma_t)=e^{-\frac{||f_t(\boldsymbol{x_1})-f_t(\boldsymbol{x_2})||_2^2}{2\sigma_t^2}}$and $S_f(\boldsymbol{x_1},\boldsymbol{x_2},\mathcal{F})=e^{-\frac{d_s(\boldsymbol{x_1},\boldsymbol{x_2},\mathcal{F})}{2\sigma_e^2}}$, where $t\in \{c,g\}$ and $d_s$ is the squared Sampson distance \cite{hartley2003multiple}. The values used for $\lambda_1$, $\lambda_2$, $\lambda_3$, $\sigma_c$, $\sigma_g$, and $\sigma_e$ are 0.15, 0.4, 0.45, 4.8, 0.25, and 0.17, respectively. $\mathcal{F}_s$ is the fundamental matrix estimated between $I_r$ and $I_s$ \cite{hartley2003multiple}. The similarity map considers appearance and geometric consistency in order to quantify the quality of correspondences.
\begin{algorithm}[t]
	\begin{algorithmic}
		\State \textbf{Input}: reference image $I_r$, source images $\{I_s\}_{s=1}^{n-1}$
		\State \textbf{Output:} Dynamic map $\mathcal{L}$, Updated reference image $\hat{I}_r$ with no dynamic objects
		\For{$s = 1 \rightarrow n-1$}
		\State Extract feature descriptors for $I_r$ and  $\{I_s\}_{s=1}^{n-1}$
		\State Compute dense correspondence map $\mathcal{N}_{r\rightarrow s}$
		\State Compute the confidence Map $\mathit{C}_{(s)}$ (Section \ref{DenseCorr})
		\EndFor
		\For{$scan \in \{down,up\}$}
		\For{$x = 1\rightarrow cols$}
		\For{$y = 1\rightarrow rows$}
		\State \makecell[l]{Find the candidate locations $P$ in $\{I_s\}_{s=1}^{n-1}$\\(Section \ref{decision})}
		\State Find $\mathcal{L}(x,y)$ using $P$ (Section \ref{decision})
		\If{$\mathcal{L}(x,y)==0$}
		\State \makecell[l]{Update $I_r(x,y)$ using patches at $\mathcal{P}$\\(Section \ref{removal})}
		\State \makecell[l]{Update $\mathcal{N}_{r\rightarrow s}(\boldsymbol{x}_{r})$ and $ \mathit{C}_{(s)}(\boldsymbol{x}_{r})$,\\ $\forall s=1,\ldots,k$ (Section \ref{update})}
		\EndIf  
		\EndFor
		\EndFor
		\EndFor
	\end{algorithmic}
	\caption{Simultaneous Detection and Removal of Dynamic Objects in Multi-view Images}
	\label{alg: CSS}
\end{algorithm}
\subsection{Simultaneous Detection and Removal of Dynamic Objects}
We rely on the coherency present in the natural images, i.e., if two patches are nearby in one image, then their nearest neighbours are likely to be close to each other in the other image of the same scene captured from a different (or same) angle. We scan the reference image in two orders: top-left to bottom-right and bottom-right to top-left. During the scan, at each pixel location, we select some candidate locations from the source images. Then, based on those candidate locations, we make a decision on whether the pixel belongs to a dynamic object or not. If the pixel belongs to a dynamic object, we update its pixel value with the pixel value of the corresponding static region, its dense correspondence map, and the similarity map. Otherwise, we move on to the next location.
\subsubsection{Decision}
\label{decision}
We select a set of candidate locations $P$ from the source images $\{I_s\}_{s=1}^{n-1}$. We compute these candidate locations similar to Generalized PatchMatch \cite{barnes2010generalized} and Kanojia \emph{et al.} \cite{kanojia2018patch}. The set of candidate locations depends on the order of scan. Let $\boldsymbol{x_r}=(\mathit{x},\mathit{y})$ be the current location in $I_r$ during the scan.
Let $\boldsymbol{\hat{x}^l_s}$, $\boldsymbol{\hat{x}^u_s}$, $\boldsymbol{\hat{x}^r_s}$, and $\boldsymbol{\hat{x}^b_s}$, be the nearest neighbour locations of the left, upper, right, and bottom of the current location $\boldsymbol{x_r}$ in the source image $I_s$, respectively. Let $\boldsymbol{x^l_s}$, $\boldsymbol{x^u_s}$, $\boldsymbol{x^r_s}$, and $\boldsymbol{x^b_s}$ be the pixel locations on the right, bottom, left, and upper of $\boldsymbol{\hat{x}^l_s}$, $\boldsymbol{\hat{x}^u_s}$, $\boldsymbol{\hat{x}^r_s}$, and $\boldsymbol{\hat{x}^b_s}$, respectively. Let $P_s$ be the set of candidate locations in the source image $I_s$ and $B_s$ be the set of their corresponding values in the similarity map $C_s$.\\
During the scan from top-left to bottom-right, $P_s = \{\boldsymbol{x^l_s},\boldsymbol{x^u_s}\}$ and $B_s = \{C_s(\mathit{x}-1,\mathit{y}),C_s(\mathit{x},\mathit{y}-1)\}$, and from bottom-right to top-left, $P_s = \{\boldsymbol{x^r_s},\boldsymbol{x^b_s}\}$ and $B_s = \{C_s(\mathit{x}+1,y),C_s(\mathit{x},\mathit{y}+1)\}$. Then, $P=\bigcup\limits_{s=1}^{n-1}P_s$ is the set of candidate locations for $\boldsymbol{x_r}$ and $B=\bigcup\limits_{s=1}^{n-1}B_s$ is the set of their corresponding values in the similarity map. Here, an entry of $B$ represents the confidence of the contender location to be the corresponding location of $\boldsymbol{x}$ in the source image. $B$ relies on the image coherency to assign weights to the contender location. It uses the similarity measure of matching of the neighbours as weights. For example, if $\boldsymbol{x_r} = (x,y)$ is the current location, then, $(x-1,y)$ is its left neighbour. Let $\boldsymbol{\hat{x}_r}= (\hat{x},\hat{y})$ be the nearest neighbour of $(x-1,y)$ in image $I_s$, then $(\hat{x}+1,\hat{y})$ is the candidate location to be the nearest neighbour of $\boldsymbol{x_r}$. The confidence of $(\hat{x}+1,\hat{y})$ to be the candidate location depends on how well $(x-1,y)$ and $\boldsymbol{\hat{x}_r}$ are matched.\\
Here, we make an assumption that the static part is exposed in majority of the images. Now, we label the current pixel location as static or dynamic. We make the decision based on the candidate locations of the current location. We apply a clustering algorithm on $P$ to obtain a set of clusters \cite{kanojia2018patch}. The distance function for the clustering algorithm is given by Eq. \ref{DB_cost}.
\begin{equation}
\label{DB_cost}
\begin{array}{ll}
\mathcal{B}(\boldsymbol{x_1},\boldsymbol{x_2}) = & \hspace{-3mm} 1- \lambda_4S_e(f_c(\boldsymbol{x_1}),f_c(\boldsymbol{x_2}),\sigma_c)\\ &-\lambda_5S_e(f_g(\boldsymbol{x_1}),f_g(\boldsymbol{x_2}),\sigma_g)
\end{array}
\end{equation}
Here, $\boldsymbol{x_1},\boldsymbol{x_2} \in P$, and $\lambda_4 = \frac{\lambda_1}{\lambda_1+\lambda_2}$, $\lambda_5 = \frac{\lambda_2}{\lambda_1+\lambda_2}$, $\sigma_c$, and $\sigma_g$ are constants. $S_e$, $f_c$, and $f_g$ are defined in section \ref{DenseCorr}. The corresponding location of the current location could be occluded in some of the source images by the same (or different) dynamic object(s). Hence, we use DBSCAN, as we do not know the number of clusters \cite{ester1996density}. Let us assume that we obtain $k$ clusters $\{A_k\}_{i=1}^k$. Let $b_k = \sum_{l}B_l$, where $B_l$ is an entry of $B$ and $l$ represents the index corresponding to the candidate locations belonging to the $k^{th}$ cluster.
\begin{equation}
(\hat{f}_c,\hat{f}_g)=(\frac{1}{b_m}\sum_{\boldsymbol{x_l} \in A_m}B_lf_c(\boldsymbol{x_l}),\frac{1}{b_m}\sum_{\boldsymbol{x_l} \in A_m}B_lf_g(\boldsymbol{x_l}))
\end{equation}
Here, $B_l$ is the entry of $B$ corresponding to $\boldsymbol{x_l}$, and $m$ is such that $b_m = \max\limits_{k} b_k$.
\begin{equation}
M(\boldsymbol{x_r})= \lambda_4S_e(f_c(\boldsymbol{x_r}),\hat{f}_c,\sigma_c)+ \lambda_5S_e(f_g(\boldsymbol{x_r}),\hat{f}_g,\sigma_g)
\end{equation}
If $M(\boldsymbol{x_r})>t_r$, then $\boldsymbol{x_r}$ belongs to the static region. Else, $\boldsymbol{x_r}$ belongs to the dynamic region. Here, $t_r$ is a constant. 
\begin{equation}
\label{eq: decision}
\mathcal{L}(\boldsymbol{x}_r) = \left\{
\begin{tabular}{ll}
1, & if $M(\boldsymbol{x_r})>t_r$\\
0, & otherwise
\end{tabular}\right.
\end{equation}
Here, $\mathcal{L}$ is the dynamic map, $0$ stands for the dynamic region and $1$ stands for the static region.
\subsubsection{Removal}
\label{removal}
If the current pixel location $\boldsymbol{x_r}$ belongs to the static region, we move on to the next location. However, if $\boldsymbol{x_r}$ belongs to a dynamic object, we update the reference image $I_r$. Let us consider that $\boldsymbol{x_r}$ belongs to a dynamic object. We can separate the candidate locations in $P$ into two parts, i.e., $\boldsymbol{x_s^a}\in A_m$ and $\boldsymbol{x_s^a}\centernot\in A_m$, where, $s\in \{1,2,\ldots,n-1\}$ and $\boldsymbol{x_s^a}$ is a candidate location. During the scan from top-left to bottom-right, $a \in \{l,u\}$ and during the scan from bottom-right to top-left $a \in \{r,b\}$ (Section \ref{decision}).\\
We update the reference image $I_r$ using candidate locations in $A_m$. Let $q_{\boldsymbol{x_s^a}}$ be an image patch of size $p\times p$ extracted from $\boldsymbol{x_s^a}$ from the source image $I_s$, where $\boldsymbol{x_s^a}\in A_m$. We extract a set of patches $\mathcal{P} = \{q_{\boldsymbol{x_s^a}}: \boldsymbol{x_s^a}\in A_m\}$ from the corresponding source images. Let $\boldsymbol{x_r^l}$, $\boldsymbol{x_r^u}$, $\boldsymbol{x_r^r}$, and $\boldsymbol{x_r^b}$ be the left, upper, right and bottom pixel locations of $\boldsymbol{x_r}$ in $I_r$, respectively. Let $q_{\boldsymbol{x_r^l}}$, $q_{\boldsymbol{x_r^u}}$, $q_{\boldsymbol{x_r^r}}$, and $q_{\boldsymbol{x_r^b}}$ be the patches of size $p\times p$ centered at $\boldsymbol{x_r^l}$, $\boldsymbol{x_r^u}$, $\boldsymbol{x_r^r}$, and $\boldsymbol{x_r^b}$, respectively. During the scan from top-left to bottom-right, $q_{\boldsymbol{x_r^l}}$ and $q_{\boldsymbol{x_r^u}}$ will be used and from bottom-right to top-left, $q_{\boldsymbol{x_r^r}}$, and $q_{\boldsymbol{x_r^b}}$ will be used.\\
First, we will discuss the scan from the top-left to bottom-right. When an image patch $q \in \mathcal{P}$ is placed at $\boldsymbol{x_r}$, let $w_q^1$ and $w_q^2$ be the overlapping region of the image patch $q$ with $q_{\boldsymbol{x_r^l}}$ and $q_{\boldsymbol{x_r^u}}$, respectively. Let $w_r^1$ and $w_r^2$ be the overlapping regions of the image patches $q_{\boldsymbol{x_r^l}}$ and $q_{\boldsymbol{x_r^u}}$ with $q$, respectively.
\begin{equation}
\begin{array}{ll}
q^\ast = & \max\limits_{q\in\mathcal{P}}\sum\limits_{i=1}^2 \big(\lambda_6\mathcal{S}_e(g_c(w_q^i),g_c(w_r^i),\sigma_c)+\\ &\lambda_7\mathcal{S}_e(g_h(w_q^i),g_h(w_r^i),\sigma_h)\big)+ \lambda_8 S_f(\boldsymbol{x_r},\boldsymbol{x_q},\mathcal{F}_q)
\end{array}
\end{equation}
Here, $g_c$ and $g_h$ are the functions which compute CIE Lab mean and rotation invariant histogram of oriented gradient (HoG) feature descriptor of the input image patch, respectively. $\boldsymbol{x_q}$ is the pixel location of the patch $q\in \mathcal{P}$ and $\mathcal{F}_q$ is the fundamental matrix between the reference image and source image $I_s$ in which the patch $q$ lies. The values used for $\lambda_6$, $\lambda_7$, $\lambda_8$, and $\sigma_h$ are 0.12, 0.36, 0.03, and 4.8 respectively.\\
We replace the patch in $I_r$ at $\boldsymbol{x_r}$ by $q^\ast$. Also, we update $f_c(\boldsymbol{x_r})$ and $f_g(\boldsymbol{x_r})$ with the CIE Lab mean feature descriptor and SIFT feature descriptor of the image patch $q^\ast$. There can be a scenario where multiple patches in $\mathcal{P}$ lie on the minima or very close to the minima. This is possible when the overlapping area is the same. However, there is a possibility that the non-overlapping area is different. Let $\hat{\mathcal{P}}$ be the set of such patches. In such a case, we replace the patch in $I_r$ at $\boldsymbol{x_r}$ by $\hat{q}$.
\begin{equation}
    \hat{q}= \min\limits_{q_{\boldsymbol{x_s^a}}\in\hat{\mathcal{P}}} \lambda_4 ||\hat{f}_c-f_c(x_s^a)||^2_2+\lambda_5||\hat{f}_g-f_g(x_s^a)||^2_2
\end{equation}
During the scan from bottom-right to top-left, we follow the same procedure except that $q_{\boldsymbol{x_r^l}}$ and $q_{\boldsymbol{x_r^u}}$ are replaced by $q_{\boldsymbol{x_r^r}}$ and $q_{\boldsymbol{x_r^b}}$, respectively and $a \in \{r,b\}$.
\subsubsection{Update}
\label{update}
After we replace the patch belonging to the dynamic object at $\boldsymbol{x_r}$ with its static counterpart, we update $\mathcal{N}_{r\rightarrow s}(\boldsymbol{x_r})$ and $C_s(\boldsymbol{x_r})$, $\forall s=1,2,\ldots,n-1$. Let,
\begin{equation}
\begin{array}{ll}
\mathcal{H}(\boldsymbol{x_r},\boldsymbol{x},\mathcal{F})= \hspace{-3mm}& \lambda_1S_e(f_c(\boldsymbol{x_r}),f_c(\boldsymbol{x}),\sigma_c)+\\ &\hspace{-6mm}\lambda_2S_e(f_g(\boldsymbol{x_r}),f_g(\boldsymbol{x}),\sigma_g)
+\lambda_3S_f(\boldsymbol{x_r},\boldsymbol{x},\mathcal{F})
\end{array}
\end{equation}
Here, $\boldsymbol{x_r}, \boldsymbol{x} \in \mathbb{R}^2$ and $\mathcal{F}$ is a fundamental matrix. Now, we have two sets of candidate locations, one that belongs to $A_m$ and the other that does not. The candidate locations were constructed in such a way that each source image contributes two candidate locations. Now, there can be three cases.\\
First, consider that only one of the two candidate locations, let us call it $\boldsymbol{x}$, of $I_s$ lies in $A_m$. Then, we have
\begin{equation}
    \begin{array}{l}
    \mathcal{N}_{r\rightarrow s}(\boldsymbol{x_r})=\boldsymbol{x}\\
    C_s(\boldsymbol{x_r})= \mathcal{H}(\boldsymbol{x_r},\boldsymbol{x},\mathcal{F}_s)
    \end{array} 
\end{equation}
Here, $\mathcal{F}_s$ is a fundamental matrix estimated between $I_r$ and $I_s$.\\
Second, consider that both the candidate locations, let us call them $\boldsymbol{x_{s_1}}$ and $\boldsymbol{x_{s_2}}$, from $I_s$ lie in $A_m$. Then, we have
\begin{equation}
    \boldsymbol{x^\ast}=\max\limits_{\boldsymbol{x}\in\{\boldsymbol{x_{s_1}},\boldsymbol{x_{s_2}}\}} \mathcal{H}(\boldsymbol{x_r},\boldsymbol{x},\mathcal{F}_s)
\end{equation}
and,
\begin{equation}
    \begin{array}{l}
    \mathcal{N}_{r\rightarrow s}(\boldsymbol{x_r})=\boldsymbol{x^\ast}\\
    C_s(\boldsymbol{x_r})= \mathcal{H}(\boldsymbol{x_r},\boldsymbol{x^\ast},\mathcal{F}_s)
    \end{array} 
\end{equation}
Third, consider that none of the candidate locations from $I_s$ lie in $A_m$. This implies that the static region corresponding to $\boldsymbol{x_r}$ in $I_s$ is occluded by a dynamic object. In this case, we cannot rely on appearance, instead we have to completely rely on geometry. Let $\boldsymbol{x_{s_1}}$ and $\boldsymbol{x_{s_2}}$ be the candidate locations picked from $I_s$ and $\boldsymbol{x_{s_1}},\boldsymbol{x_{s_2}}\centernot\in A_m$ . During the scan from top-left to bottom-right, $X = \{\boldsymbol{x_{s_1}},\boldsymbol{x_{s_2}},\boldsymbol{\hat{x}_s^l}, \boldsymbol{\hat{x}_s^u}\}$, and from bottom-right to top-left, $X = \{\boldsymbol{x_{s_1}},\boldsymbol{x_{s_2}},\boldsymbol{\hat{x}_s^r}, \boldsymbol{\hat{x}_s^b}\}$. Then,
\begin{equation}
    \boldsymbol{x^\ast}=\max\limits_{\boldsymbol{x_s}\in X} S_f(\boldsymbol{x_r},\boldsymbol{x_s},\mathcal{F}_s)
\end{equation}
and,
\begin{equation}
    \begin{array}{l}
    \mathcal{N}_{r\rightarrow s}(\boldsymbol{x_r})=\boldsymbol{x^\ast}\\
    C_s(\boldsymbol{x_r})= \mathcal{H}(\boldsymbol{x_r},\boldsymbol{x^\ast},\mathcal{F}_s)
    \end{array} 
\end{equation}
\begin{figure}[t]
	\centering
	\includegraphics[width=0.85\linewidth]{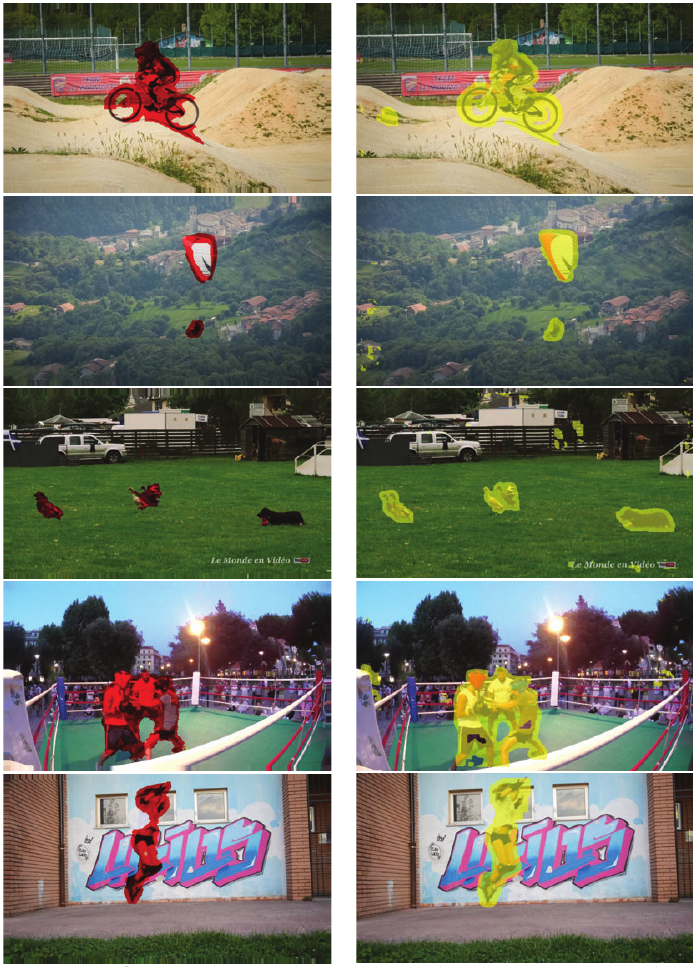}
	\caption{The figure shows the comparsion between the detection results obtained in Kanojia \emph{et al} \cite{kanojia2018patch} and using the proposed approach. The left column shows the results obtained in Kanojia \emph{et al.} \cite{kanojia2018patch} and the right column shows the results obtained using the proposed approach.}
	\label{detection results}
\end{figure}
\begin{figure}[t]
	\centering
	\includegraphics[width=0.95\linewidth]{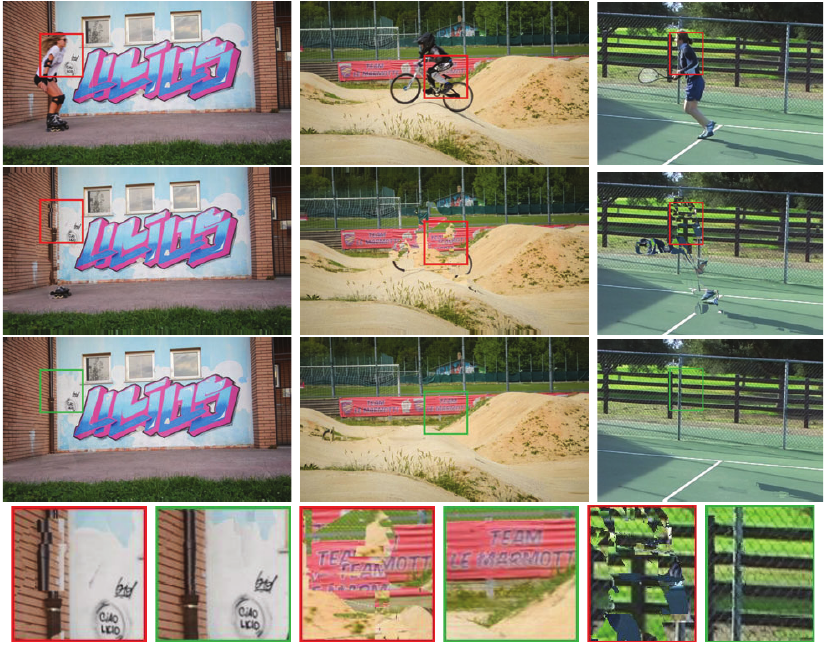}
	\caption{The figure shows the comparison between the object removal results obtained using Kanojia \emph{et al.} \cite{kanojia2018patch} and the results obtained using the proposed approach. The first row shows the reference images of some of the datasets. The second row shows the results obtained using the approach by Kanojia \emph{et al.} \cite{kanojia2018patch}. The third shows the results obtained using the proposed approach. }
	\label{fig:Comparison 1}
\end{figure}
These matches will have low similarity value. Hence, they will not affect the selection of the most confident cluster, i.e., $A_m$ for the next location. This will continue during the scan until the dynamic region completely passes in that source image. The geometry helps the match to slide over the dynamic region while keeping it close to the occluded static counterpart of $\boldsymbol{x_r}$ in that source image. The reason behind including $\{\boldsymbol{\hat{x}_s^l}, \boldsymbol{\hat{x}_s^u}\}$ and $\{\boldsymbol{\hat{x}_s^r}, \boldsymbol{\hat{x}_s^b}\}$ is the image warping due to wide baseline. The corresponding location of $\boldsymbol{x_r}$ may not always increment in the source image as we proceed in the scan.\\

\begin{figure*}[htbp]
	\centering
	\includegraphics[width=0.98\linewidth]{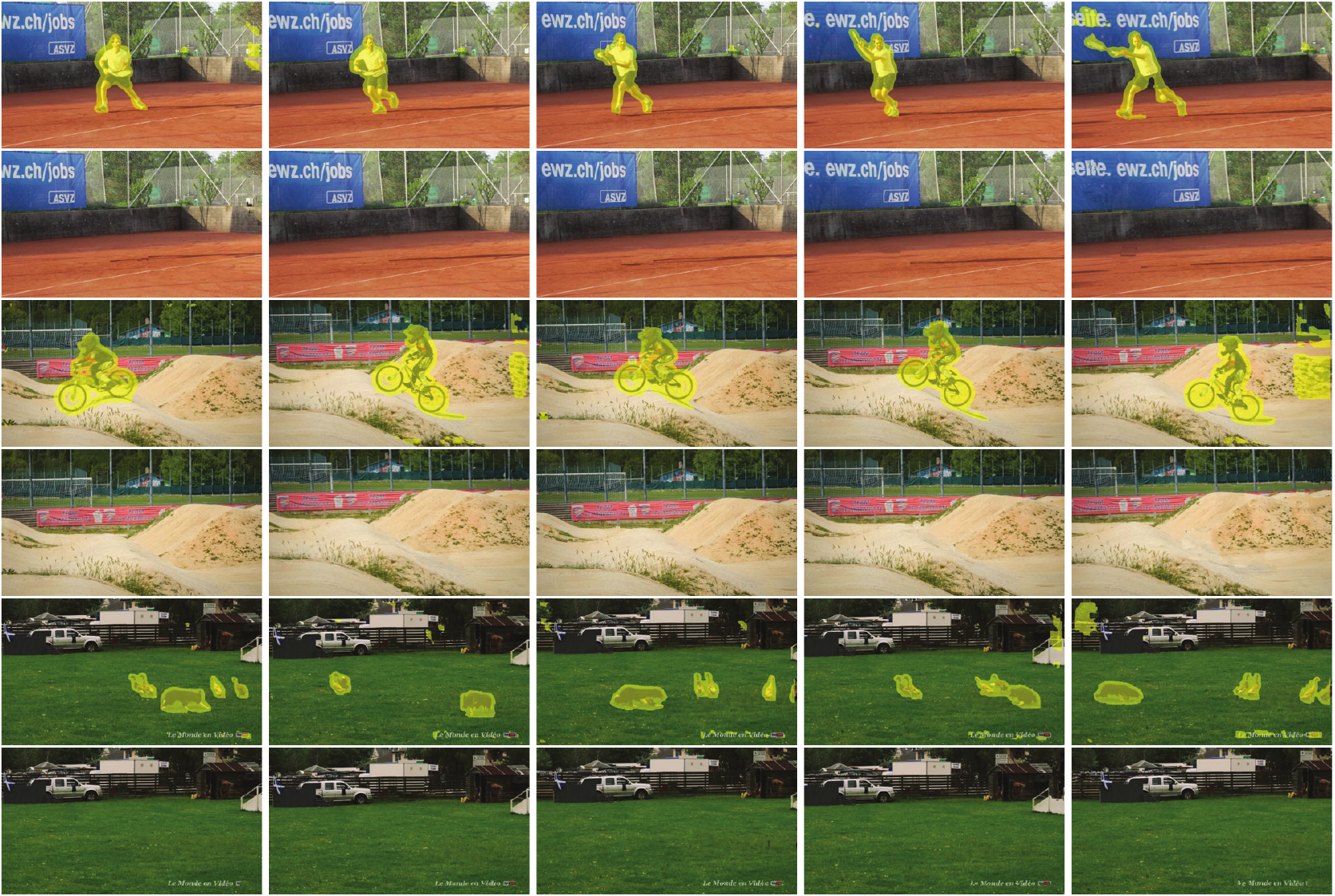}
	\caption{The figure shows the detection and the removal results obtained on the multi-view datasets extracted from DAVIS dataset \cite{Perazzi2016} using the proposed approach. It can be observed in each image set that the dynamic objects has been replaced by the static regions which were occluded by them.}
	\label{fig:results set}
\end{figure*}
\begin{table}[h]
	\centering
	\begin{tabular}{|c|c|c|c|}
		\hline
		Dataset & \makecell{Dafni \emph{et al.} \\ \cite{dafni2017detecting}} & \makecell{Kanojia \\ \emph{et al.} \cite{kanojia2018patch}} & Ours \\ \hline
		Skateboard &$0.42$ &$0.5$ &$0.67$\\
		Basketball&$0.47$ &$0.51$ &$0.47$\\
		Climbing &$0.13$ &$0.34$ &$0.28$ \\
		Playground &$0.32$ &$0.36$ &$0.43$ \\
		Toy ball &$0.6$ &$0.44$ &$0.31$ \\ \hline
	\end{tabular}
	\vspace{1mm}
	\caption{The table shows the comparison of the dynamic object detection results between Dafni \emph{et al.} \cite{dafni2017detecting}, Kanojia \emph{et al.} \cite{kanojia2018patch}, and the proposed approach on the CrowdCam image sets used in \cite{dafni2017detecting} in terms of Jaccard index. We obtain better/comparable results with the state-of-the-art even when we are not only focusing on the detection but also the removal of the dynamic objects.}
	\label{table: Jaccard 1}
\end{table}
\begin{figure}[t]
	\centering
	\includegraphics[width=0.9\linewidth]{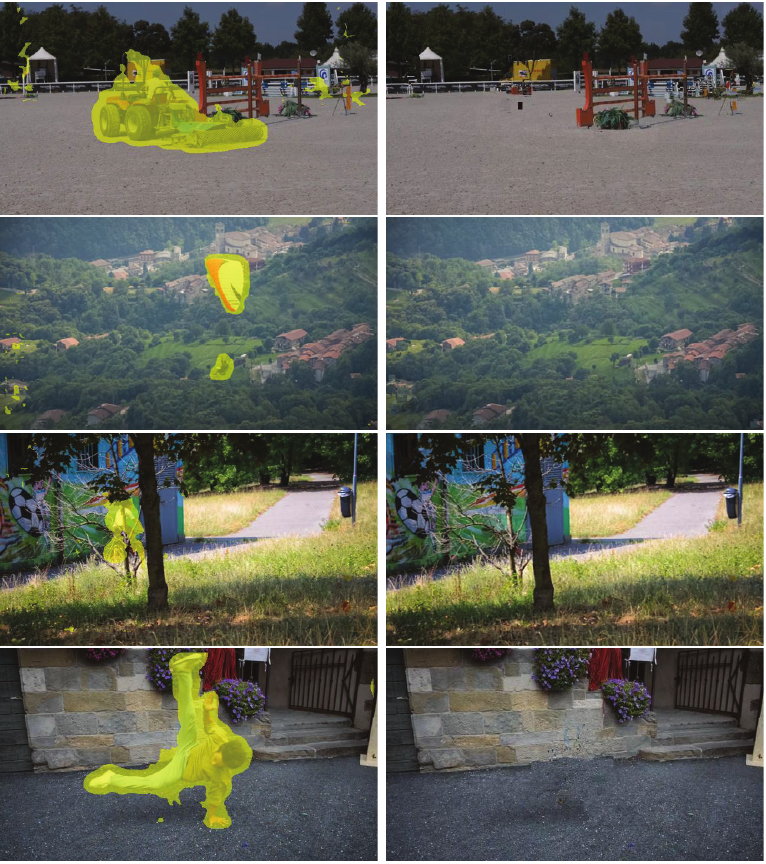}
	\caption{The figure shows the dynamic object detection and removal results obtained on the reference image of four multi-view image sets extracted from Davis dataset \cite{Perazzi2016} using the proposed approach.\vspace{-1em}}
	\label{fig:results 1}
\end{figure}
\section{Results and Discussion}
\label{Results and Discussions}
\begin{figure}[t]
	\centering
	\includegraphics[width=0.9\linewidth]{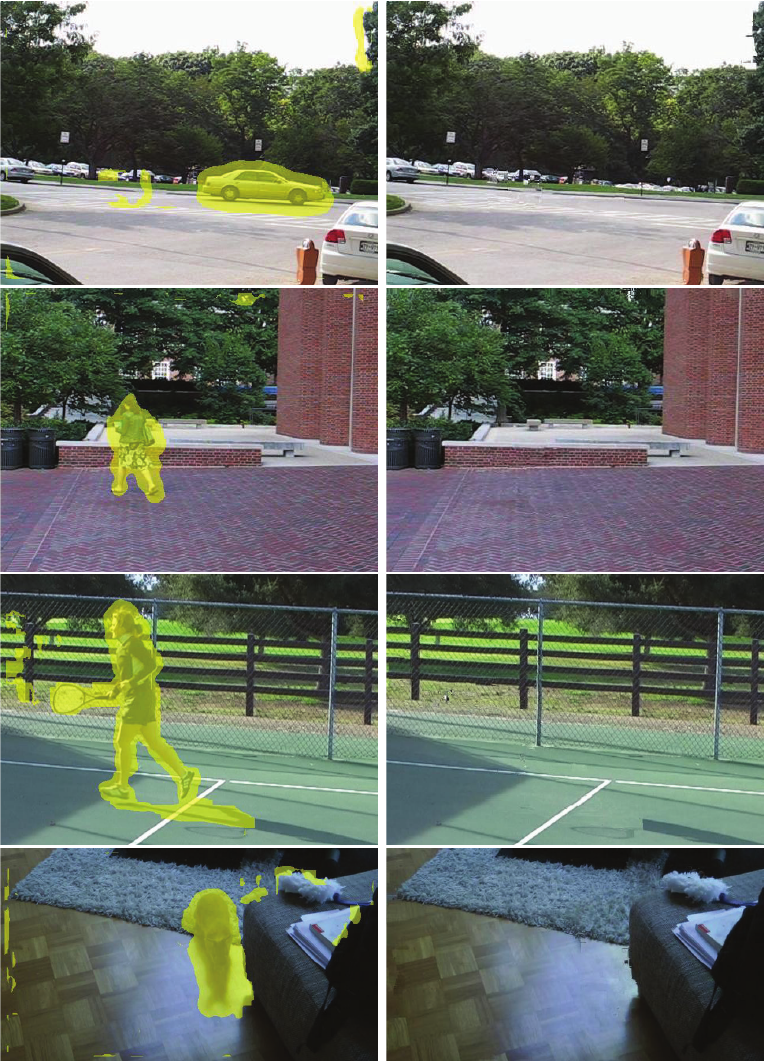}
	\caption{The figure shows the dynamic object detection and removal results obtained on the reference image of four multi-view image sets extracted from Freiburg Berkeley Motion Segmentation Dataset \cite{brox2010freiburg} using the proposed approach.\vspace{-1em}}
	\label{fig:results 2}
\end{figure}

\noindent{\bf Dataset.} As, a dedicated dataset of multi-view images with dynamic objects is not publicly available, we constructed the dataset for this work as follows. We selected some scenes from the DAVIS dataset \cite{Perazzi2016}. We sampled frames at an interval of around 6-10 frames to create image sets with 5-7 images in each set. Similarly, we extracted some multi-view sets from Freiburg Berkeley Motion Segmentation dataset \cite{brox2010freiburg}. We also used the skateboard, basketball, climbing, playground, and toyball datasets used in Dafni \emph{et at.} \cite{dafni2017detecting}, and tennis dataset used in \cite{brox2011large}. We have used the VLFeat implementation of
dense SIFT feature descriptors in all our experiments \cite{vedaldi2010vlfeat}.\\
\noindent{\bf Results.} We applied the proposed algorithm on the image sets from the prepared dataset to obtain the results. In Fig. \ref{fig: iteration}, we show the progress of the detection and the removal of the dynamic object at certain intervals during the first scan of the algorithm. Fig. \ref{fig: iteration}(a) and \ref{fig: iteration}(b) show the reference image and the source images, respectively. The first and the second row of Fig. \ref{fig: iteration}(c) show the detection and the removal of the dynamic object in the reference image at certain intervals during the first scan, respectively. In each column of Fig. \ref{fig: iteration}(c), the detection and the removal results are shown for the same iteration. It can be seen that the dynamic objects are being detected and removed, simultaneously. In the third row of Fig. \ref{fig: iteration}(c), it can be observed that the text and the symbols which were occluded by the dynamic object have been properly filled in the dynamic region. It can be observed that the text and the symbols which are getting updated in the reference image are consistent with the corresponding regions in the sources images in which those static regions are not occluded. This example shows the efficiency of the algorithm in transferring the static regions from the source images into the reference image without any artifacts. In general, the proposed approach requires multiple scans of the reference image to arrive at an artifact-free transfer of the static regions from the source images to the reference image.\\
The algorithm proposed in \cite{kanojia2018patch} mainly deals with the detection of the dynamic objects present in the images of a scene. They showed some preliminary results of their proposed algorithm on the removal of the dynamic objects when the scene is captured using a static camera. However, such assumptions are not valid when the images are captured using a hand-held camera. In Fig. \ref{detection results}, we compare the dynamic object detection results obtained in Kanojia \emph{et al.} \cite{kanojia2018patch} with the results obtained using the proposed approach. It can be observed that we obtain better coverage over the dynamic object which is very crucial for the artifact-free removal of the dynamic objects. Table \ref{table: Jaccard 1} shows the comparison of the dynamic object detection results obtained in Dafni \emph{et al.} \cite{dafni2017detecting} and Kanojia \emph{et al.} \cite{kanojia2018patch} with our results on the datasets used in \cite{dafni2017detecting} in terms of Jaccard index used in \cite{dafni2017detecting}. We obtain better/comparable results with the state-of-the-art even when we are not only focusing on the detection but also the removal of the dynamic objects. In Fig. \ref{fig:Comparison 1}, we compare the dynamic object removal results obtained on image sets captured using hand-held cameras using the approach by Kanojia \emph{et al.}\cite{kanojia2018patch} with the proposed approach. The first row shows the reference images of some of the datasets. The second row shows the results obtained using the approach by Kanojia \emph{et al.} \cite{kanojia2018patch}. The third shows the results obtained using the proposed approach. It can be seen that the proposed algorithm performs much better in comparison to \cite{kanojia2018patch}. In all our experiments, the threshold used in DBSCAN and the threshold $t_r$ used in Eq. \ref{eq: decision} range beween 0.15 to 0.8. In Fig. \ref{fig:results set}, we show the detection and the removal results obtained on the multi-view datasets extracted from DAVIS dataset \cite{Perazzi2016} using the proposed approach. It can be observed in each image set that dynamic objects has been replaced by static regions which were occluded by them.\\
In Fig. \ref{fig:results 1}, we show the detection and the removal results obtained on the reference images of four multi-view image sets extracted from Davis dataset \cite{Perazzi2016} using the proposed approach. In Fig. \ref{fig:results 2}, we show the detection and the removal results obtained on the reference images of four multi-view image sets extracted from Freiburg Berkeley Motion Segmentation Dataset \cite{brox2010freiburg} using the proposed approach. The results for the complete set for the image set shown in Fig. \ref{fig:results 1} and \ref{fig:results 2} are provided in the supplementary material.\\
 The previous learning-based image completion works used a single image as the input \cite{iizuka2017globally,yu2018generative} and the networks were trained on datasets like Places2 \cite{zhou2018places}. On the other hand, the proposed approach utilizes multiple images to not only fill the dynamic objects but also to detect them. Hence, a fair comparison is not plausible. However, just for reference, we have provided some comparisons with the learning-based single image inpainting methods in the supplementary material. We have also provided some more qualitative and quantitative results in the supplementary material.

\section{Conclusion and Future Work}
\label{Conclusion and Future Work}
We have designed a novel framework which detects the moving objects present in the multi-view images while simultaneously removing them. We replace the moving objects with the static regions which are occluded by them. We do not rely on the quality of correspondences obtained on the dynamic objects. However, the quality of detection and removal depends on the quality of correspondences obtained in the static region. We exploit image coherency and epipolar geometry to detect and remove the dynamic objects. Also, we do not take any user assistance. Our algorithm does not involve 3D reconstruction of the scene which in itself needs handling of the dynamic objects. We show that we achieve an artifact-free transfer of static regions from the source images to the reference image for several complex real-world scenes.\\

\noindent\textbf{Acknowledgments.} Gagan Kanojia was supported by TCS Research Fellowship. Shanmuganathan Raman was supported by SERB Core Research Grant and Imprint 2 Grant.
{\small
\bibliographystyle{ieee}
\bibliography{egbib}
}

\end{document}